\documentclass[
]{ceurart}

\usepackage{listings}
\usepackage{multicol}
\usepackage{multirow}
\usepackage{booktabs}
\usepackage{graphicx}

\begin{document}

\copyrightyear{2025}
\copyrightclause{Copyright for this paper by its authors.
  Use permitted under Creative Commons License Attribution 4.0
  International (CC BY 4.0).}

\conference{4DMR@IJCAI25: International IJCAI Workshop on 1st Challenge and Workshop for 4D Micro-Expression Recognition for Mind Reading, August 29, 2025, Guangzhou, China}

\title{DIANet: A Phase-Aware Dual-Stream Network for Micro-Expression Recognition via Dynamic Images}

\author[]{Vu Tram Anh Khuong}
\author[]{Luu Tu Nguyen}
\author[]{Thi Bich Phuong Man}
\author[]{Thanh Ha Le}
\author[]{Thi Duyen Ngo}[
email=duyennt@vnu.edu.vn,
]
\cormark[1]

\address[]{Faculty of Information Technology, VNU University of Engineering and Technology, Hanoi, Vietnam}

\cortext[1]{Corresponding author}

\begin{abstract}
Micro-expressions are brief, involuntary facial movements that typically last less than half a second and often reveal genuine emotions. Accurately recognizing these subtle expressions is critical for applications in psychology, security, and behavioral analysis. However, micro-expression recognition (MER) remains a challenging task due to the subtle and transient nature of facial cues and the limited availability of annotated data. While dynamic image (DI) representations have been introduced to summarize temporal motion into a single frame, conventional DI-based methods often overlook the distinct characteristics of different temporal phases within a micro-expression. To address this issue, this paper proposes a novel dual-stream framework, DIANet, which leverages phase-aware dynamic images - one encoding the onset-to-apex phase and the other capturing the apex-to-offset phase. Each stream is processed by a dedicated convolutional neural network, and a cross-attention fusion module is employed to adaptively integrate features from both streams based on their contextual relevance. Extensive experiments conducted on three benchmark MER datasets (CASME-II, SAMM, and MMEW) demonstrate that the proposed method consistently outperforms conventional single-phase DI-based approaches. The results highlight the importance of modeling temporal phase information explicitly and suggest a promising direction for advancing MER.
\end{abstract}


\maketitle

\section{Introduction}
\label{s: intro}

Micro-expressions (MEs) are brief, involuntary facial movements that typically last less than 0.5 seconds and reveal genuine emotions that an individual may attempt to conceal. Unlike macro-expressions, MEs are subtle and transient, often involving only localized muscle activations. These characteristics make MEs difficult to detect by human observers and pose significant challenges for automatic micro-expression recognition (MER) systems. Nevertheless, accurate MER has substantial value in various real-world applications such as lie detection, security screening, psychotherapy, and human–computer interaction~\cite{Bhushan2015, yan2013fast, polikovsky2010detection}.

Early research in MER primarily relied on handcrafted features designed to capture spatial texture (e.g., Local Binary Patterns from Three Orthogonal Planes, LBP-TOP~\cite{li2013spontaneous}) or optical strain~\cite{liong2014optical}, combined with temporal analysis through optical flow~\cite{Pfister2011}. While effective in controlled settings, these methods often struggle with robustness under cross-subject variation or real-world noise and require extensive feature engineering. In recent years, the emergence of deep learning and motion-based representations has led to more compact and learnable frameworks for MER. Among them, the dynamic image (DI) representation has gained popularity. Originally proposed in action recognition, DI summarizes the temporal evolution of motion into a single image using rank pooling~\cite{bilen2016dynamic}. In the context of MER, DI-based models such as LEARNet~\cite{li2022deeplearningmicroexpressionrecognition} demonstrate that DI can effectively encode subtle motion signals in a suitable form for convolutional neural networks. Extensions of this idea include Active Image~\cite{verma2020non}, which highlights salient motion regions, and Affective Motion Imaging~\cite{verma2021affectivenetaffectivemotionfeaturelearningfor}, which emphasizes emotion-relevant motion features. These methods show that DI offers an efficient way to encode facial motion into a single frame.

However, existing DI-based approaches typically treat the entire sequence as a whole, ignoring the distinct temporal phases of MEs (i.e., onset-apex and apex-offset). This holistic modeling overlooks the asymmetric rise-and-fall dynamics of facial motion. Incorporating phase-specific representations could better capture these patterns, yet remains underexplored in the literature. Phase-aware modeling has been explored in a few recent works. For example, Liong \textit{et al.}~\cite{liong2018less} emphasized the onset–apex segment by computing optical flow between two key frames, while Zhang \textit{et al.}~\cite{zhang2021facial} proposed apex-guided representation learning. However, these approaches typically rely on hand-crafted segmentation or optical flow and do not fully integrate phase modeling into a learnable deep framework. Moreover, none of them explicitly leverage DI to encode motion separately within each phase.

To address this gap, this paper proposes DIANet, the first dual-stream MER framework that uses phase-specific dynamic images to separately model the onset–apex and apex–offset intervals. By processing these phases in parallel, the model captures both the rising and falling motion patterns of MEs - subtle dynamics often overlooked in single-stream approaches. This phase-aware design enhances representation learning and leads to more accurate recognition of subtle and transient expressions. To the best of our knowledge, this is the first work to explicitly integrate phase-specific DI representations into an end-to-end framework for micro-expression recognition.

The remainder of the paper is organized as follows. 
Section~\ref{s: method} describes the proposed DIANet architecture in detail. Next, section~\ref{s:experiments and results} describes the experimental setup and reports results on CASME-II, SAMM, and MMEW. Finally, section~\ref{s: conclusion} concludes the paper and discusses future directions.

\section{Proposed method}
\label{s: method}
\begin{figure*}[ht]
    \centering
    \includegraphics[width=\linewidth]{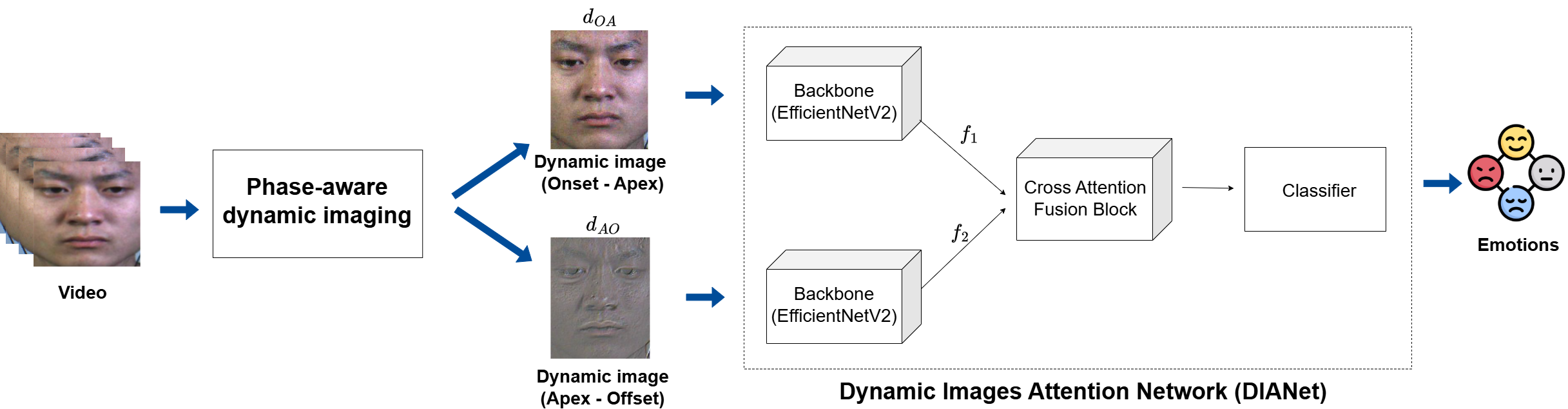}
    \caption{The workflow of our proposed dual-stream dynamic imaging for micro-expression recognition}
    \label{fig:pipeline}
\end{figure*}

This paper introduces a dual-stream dynamic imaging (DI) framework to address the limitations of conventional DI-based MER. By decomposing facial motion into two distinct temporal phases (i.e., onset-apex or apex-offset), this paper enhances motion representation while suppressing noise. The overall workflow is illustrated in Fig.~\ref{fig:pipeline}.

\begin{figure*}[ht]
    \centering
    \includegraphics[width=0.7\linewidth]{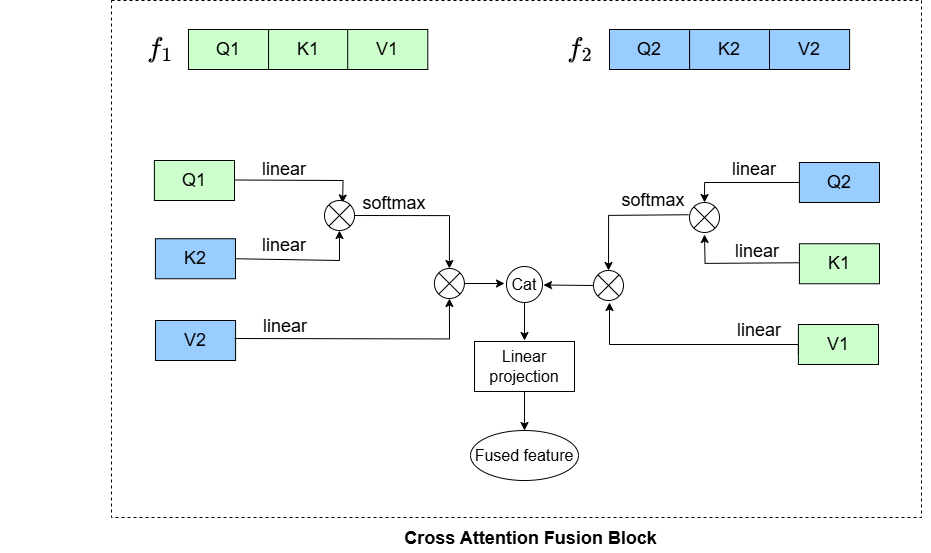}
    \caption{The cross-attention fusion block in DIANet. Inputs $f1$ and $f2$ are feature vectors obtained from the outputs of two EfficientNetV2 backbones corresponding to the onset–apex and apex–offset streams, respectively. Each is projected into query, key, and value spaces, and mutual attention is computed to produce a fused representation}
    \label{fig: attention}
\end{figure*}

\subsection{Phase-aware dynamic imaging}
Dynamic image is a technique that summarizes a video sequence into a single image, encoding both spatial and temporal information into a static representation. It is commonly constructed using Approximate Rank Pooling (ARP)~\cite{bilen2016dynamic}, where each frame in a sequence is projected into a feature space and combined with a weight proportional to its temporal position. Formally, given a sequence of $T$ consecutive frames ${F_1, F_2, \dots, F_T}$ and corresponding features $\psi(F_t)$, the dynamic image $d^*$ is computed as follows:

\begin{equation}
\label{eq:DI}
d^* = \sum_{t=1}^{T} \alpha_t \psi(F_t), \quad \text{with} \quad \alpha_t = 2t - T - 1.
\end{equation}

where $\psi(F_t) = F_t$, since raw pixel values are directly employed as frame-level features in each color channel.

The standard ARP assigns higher weights to later frames, assuming motion accumulates over time - a valid assumption for full-length actions, but not always in the case of micro-expressions. Micro-expressions typically follow a bell-shaped intensity curve, with motion rising to a peak (apex) and then declining. Applying ARP over the full sequence skews the resulting dynamic image toward the offset phase, underrepresenting early-phase motion where key emotional cues often occur. 

To overcome this limitation, this paper proposes a dual-phase dynamic imaging strategy that separately models the onset–apex and apex–offset phases using a tailored ARP formulation. Each segment is used to construct a dedicated phase-specific DI.

\begin{itemize}
\item \textbf{DI-Onset:} Captures the rising motion from onset to apex using the standard ARP with increasing weights, as shown in~Eq.\ref{eq:DI1}. This emphasizes frames closer to the apex, where expressive intensity typically peaks.
\begin{equation}
\label{eq:DI1}
d_{OA} = \sum_{t=1}^{T} \alpha_t \psi(F_t), \quad \text{with} \quad \alpha_t = 2t - T - 1.
\end{equation}
where $t$ indexes the frames in the onset-apex segment of length $T$.

\item \textbf{DI-Offset:} Encodes the declining motion from apex to offset using reversed ARP with decreasing weights defined as follows:
\begin{equation}
\label{eq:DI2}
d_{AO} = \sum_{t=1}^{T} \alpha_t' \psi(F_t), \quad \text{with} \quad \alpha_t' = T + 1 - 2t.
\end{equation}
Here, $t$ indexes the frames in the apex–offset segment of length $T$, and higher weights are assigned to frames near the apex to highlight the transition back to a neutral state.
\end{itemize}

This dual-phase formulation has two key advantages. First, it preserves the distinct motion characteristics of each phase, allowing the model to learn more fine-grained, phase-specific features. Second, it mitigates the temporal bias inherent in single-DI representations by balancing the representation across both phases. In summary, by applying ARP in a phase-aware manner, our dual-phase dynamic imaging method captures both the escalation and de-escalation of facial expressions, providing a more comprehensive and balanced motion representation for MER.

\subsection{Dynamic Images Attention Network (DIANet)}
\label{s: classification}

To fully exploit the asymmetric and complementary motion patterns present in micro-expression sequences, this paper proposes a novel dual-stream classification architecture called Dynamic Images Attention Network (DIANet) (as illustrated in Fig~\ref{fig:pipeline}). Unlike conventional approaches that rely on a single dynamic image over the entire sequence, DIANet introduces a phase-aware formulation by explicitly modeling the two key motion intervals: onset–apex (rising phase) and apex–offset (fading phase). Each phase is encoded into a dynamic image, referred to as DI-Onset and DI-Offset, which are processed independently in parallel streams.

Each stream in DIANet employs a shared backbone based on a modified EfficientNetV2\cite{tan2021efficientnetv2} architecture. The classification head is removed to retain only high-level feature extraction, ensuring consistency and parameter efficiency across both branches. This backbone was selected based on its superior performance in the ablation study (see section~\ref{s:ablation}), offering a strong balance between compactness and representational power - crucial for handling subtle motion in limited micro-expression data. 

For each input stream, the backbone outputs a 512-dimensional feature vector: $f1$ for the onset–apex phase and $f2$ for the apex–offset phase. To bridge the semantic gap and promote information exchange between the two phases, DIANet introduces a novel cross-attention fusion module, as illustrated in Fig.~\ref{fig: attention}. This module enables bidirectional interaction between DI-Onset and DI-Offset features through learnable query-key-value projections. Each feature vector attends to the other, allowing the model to focus on salient motion cues across both phases. The resulting attention-enhanced features are concatenated and passed through a projection layer to form a unified representation.

The fused feature is then passed through a lightweight multilayer perceptron (MLP) classifier, consisting of two fully connected layers with ReLU activation and dropout regularization, to predict the final emotion class. This architectural design balances expressiveness and efficiency, making it suitable for micro-expression recognition under limited data conditions.

\subsection{Loss function}
\label{s:loss}
To enhance representational coherence between the two temporal phases, this paper introduces a consistency regularization term that encourages alignment between the features learned from the onset–apex and apex–offset streams. Specifically, a cosine-based consistency loss is applied to penalize discrepancies between the corresponding feature vectors. To the best of our knowledge, this is the first application of cross-phase regularization in the context of micro-expression recognition using dynamic image representations.

The consistency loss is defined as the average cosine dissimilarity across a mini-batch:
\begin{equation}
\label{eq:cons_loss}
\mathcal{L}_{\text{cons}} = \frac{1}{N} \sum_{i=1}^{N} \left(1 - \cos\left(f_1^{(i)}, f_2^{(i)}\right)\right),
\end{equation}
where \( f_1^{(i)} \) and \( f_2^{(i)} \) denote the feature vectors from the DI-Onset and DI-Offset streams for the \(i\)-th sample, and \( N \) is the batch size.

The total training objective combines the standard cross-entropy classification loss \( \mathcal{L}_{\text{CE}} \) with the proposed consistency loss:
\begin{equation}
\label{eq:total_loss}
\mathcal{L}_{\text{total}} = \mathcal{L}_{\text{CE}} + \lambda \cdot \mathcal{L}_{\text{cons}},
\end{equation}
where \( \lambda \) is a regularization coefficient that balances classification accuracy and cross-phase feature alignment. This joint objective promotes both discriminative learning and temporal consistency, leading to more robust and generalizable micro-expression recognition.

\section{Experiments and Results}
\label{s:experiments and results}


\subsection{Datasets}

To evaluate the effectiveness of the proposed method, experiments are conducted on three publicly available benchmark datasets commonly used in micro-expression recognition:

\begin{itemize}
\item \text{CASME-II}\cite{casme}: This dataset consists of 255 spontaneous micro-expression samples from 26 subjects. Recordings were captured under laboratory conditions at a high frame rate of 200 fps, allowing precise temporal localization of facial movements. All samples were manually annotated with emotion labels and apex frames.

\item \text{SAMM}\cite{samm}: The SAMM dataset contains 159 high-resolution micro-expression samples from 32 participants. Similar to CASME-II, the recordings were captured at 200 fps in a controlled environment. Emotion annotations were provided by expert coders based on FACS (Facial Action Coding System) criteria.

\item \text{MMEW}~\cite{ben2021video}: The MMEW dataset includes 300 spontaneous micro-expression samples collected in more naturalistic settings. Videos were recorded at 90 fps and annotated with six emotion categories. MMEW introduces more variation in head pose and lighting, making it a challenging benchmark for generalization.
\end{itemize}

To ensure balanced evaluation, emotion classes with fewer than 10 samples are excluded. After filtering, CASME-II is reduced to five categories (i.e., disgust, happiness, surprise, repression, others), SAMM to five categories (i.e., happiness, surprise, anger, contempt, others) and MMEW to six categories (i.e., happiness, sadness, fear, others, surprise, disgust). To improve generalization and address data scarcity, standard data augmentation techniques are employed during training. These include horizontal flipping and random in-plane rotations within the range of ±10 degrees. All video sequences are temporally normalized and resized to a fixed spatial resolution before being processed into dynamic image representations.

\subsection{Evaluation Protocol}
The evaluation follows the Leave-One-Subject-Out (LOSO) cross-validation protocol, which is commonly used in the MER field to assess subject-independent generalization. In each iteration, the samples from one subject are reserved for testing, while the remaining data are used for training.






\subsection{Evaluation Metric}

Model performance is evaluated using overall Accuracy (Acc), which reflects the proportion of correctly classified samples over the total number of samples, as shown in Eq~\ref{eq: acc}:

\begin{equation}
\label{eq: acc}
\text{Acc} = \frac{\text{Number of Correct Predictions}}{\text{Total Number of Samples}}
\end{equation}


\subsection{Implementation Details}
All experiments are conducted using PyTorch. Input dynamic images are resized to 224×224 and normalized. Models are trained for 50 epochs using the Adam optimizer with a learning rate of $10^{-4}$ and a batch size of 32. Early stopping is applied based on validation loss. The training objective includes the consistency loss described in the section~\ref{s:loss} to encourage alignment between phase-specific feature representations.

\subsection{Results}
\label{s: results}

Table~\ref{tab:sota} presents a comparative analysis of the proposed method (DIANet) against recent state-of-the-art approaches that utilize dynamic images or DI-inspired representations. The evaluation is conducted across three widely used micro-expression datasets: CASME-II, SAMM, and MMEW. 

Our proposed method DIANet achieves the highest accuracy on two of the three datasets: 68.89\% on SAMM and 64.24\% on MMEW, demonstrating its robustness across both high-resolution and naturalistic micro-expression scenarios. These results highlight the advantage of phase-aware modeling in capturing localized motion patterns that are often overlooked in single-phase or holistic DI-based approaches. On CASME-II dataset, DIANet achieves a competitive accuracy of 70.00\%, demonstrating a performance comparable to recent advanced methods such as GEME~\cite{nie2021geme} (75.20\%) and FDCN~\cite{tang2023novel} (73.09\%). GEME leverages gender information as an auxiliary input to capture identity-dependent dynamics, but at the cost of increased model complexity and reliance on demographic metadata, which may raise practical and ethical concerns. FDCN combines dynamic images with optical flow to enrich motion features, yet its multi-stream architecture introduces additional computational overhead and modality alignment challenges.implementation overhead. In contrast, DIANet relies solely on phase-specific dynamic images derived from the original video sequence, requiring no auxiliary labels or multi-modal fusion. By explicitly separating motion into the onset-apex and apex-offset phases, DIANet learns more fine-grained, phase-sensitive features that are often lost in conventional DI-based approaches. This design not only simplifies the model architecture but also improves interpretability and portability across datasets.

\begin{table*}[ht]
\centering
\caption{Comparison with DI-based state-of-the-art MER methods (accuracy \%)}
\label{tab:sota}
\begin{tabular}{cccccc}
\toprule
\textbf{Year} & \textbf{Input} & \textbf{Method} & \textbf{CASME-II} & \textbf{SAMM} & \textbf{MMEW} \\
\midrule
2019 & DI & LEARNet*~\cite{verma2019learnet} & 56.67 & 59.26 & 48.96 \\
2020 & Active image & OrigiNet~\cite{verma2020non} & 62.09 & 34.89 & - \\
2021 & Affective-motion image & AffectiveNet~\cite{verma2021affectivenetaffectivemotionfeaturelearningfor} & 46.49 & 58.12 & - \\
2021 & DI + gender & GEME~\cite{nie2021geme} & \textbf{75.20} & 55.88 & - \\

2023 & DI + Optical flow & FDCN~\cite{tang2023novel} & 73.09 & 58.07 & - \\
\text{-} & \textbf{DI-Onset + DI-Offset} & \textbf{DIANet (ours)} & 70.00 & \textbf{68.89} & \textbf{64.24} \\
\bottomrule
\end{tabular}

\vspace{5pt} 
\small{ * This results are from our re-implementation with the model provided by the author}
\end{table*}

Compared to LEARNet\cite{verma2019learnet}, a widely used DI-based baseline, DIANet yields substantial improvements: +11.22\% on CASME-II, +9.63\% on SAMM, and +15.28\% on MMEW. These gains confirm that explicitly modeling phase transitions enhances the model's ability to capture expressive facial motion. Other DI-motivated methods, such as OrigiNet\cite{verma2020non} and AffectiveNet~\cite{verma2021affectivenetaffectivemotionfeaturelearningfor}, also fall behind in generalization, with particularly low performance on SAMM (e.g., 34.89\%), suggesting that simple enhancements over DI are insufficient for cross-dataset robustness.

In summary, the consistent performance of DIANet across diverse datasets without reliance on handcrafted rules, demographic information, or multi-modal fusion, demonstrates its effectiveness and generalization. The results underscore the importance of phase-aware modeling as a lightweight yet powerful strategy for advancing micro-expression recognition in both constrained and real-world environments.

\subsection{Ablation study}
\label{s:ablation}
\subsubsection{Backbone selection for DIANet}
The performance of DIANet is influenced not only by its dual-stream, phase-aware design but also by the feature extraction backbone used in each stream. To investigate the impact of backbone architecture, we evaluate four representative convolutional backbones: ResNet18~\cite{he2016deep}, EfficientNetV2~\cite{tan2021efficientnetv2}, ConvNeXt~\cite{liu2022convnet}, and MobileViT~\cite{mehta2022mobilevit}. These models are selected based on their widespread use, architectural diversity, and relevance to tasks involving subtle motion recognition under data constraints. Table~\ref{tab:backbone} illustrates the performance of DIANet using different backbone architectures across three benchmark datasets.

\begin{table*}[h]
\caption{Comparison of backbone performance (accuracy \%) on DIANet across three datasets}
    \centering
    \begin{tabular}{cccc}
    \hline
    \toprule
      \textbf{Backbone} &\textbf{ CASME-II} & \textbf{SAMM} & \textbf{MMEW} \\
      \midrule
        Resnet18~\cite{he2016deep} & 61.67 & 64.44 & \textbf{64.94} \\
        EfficientNetV2~\cite{tan2021efficientnetv2} & \textbf{70.00} & \textbf{68.89} & 64.24 \\
        ConvNeXt~\cite{liu2022convnet} & 59.17&58.52 &60.45\\
        MobileViT~\cite{mehta2022mobilevit} &62.08 &60.74 &62.08 \\
        \bottomrule
         
    \end{tabular}

    \label{tab:backbone}
\end{table*}

All backbones are integrated into the same DIANet framework and trained under identical experimental settings. Each model processes a pair of dynamic images corresponding to the onset–apex and apex–offset phases, and outputs are fused via the same cross-attention mechanism. The goal is to isolate the effect of the backbone itself on performance across three micro-expression datasets.

\begin{itemize}
    \item ResNet18 serves as a strong and widely adopted baseline. It provides a good trade-off between depth and computational cost, making it suitable for MER tasks with limited training data. On MMEW, it surprisingly achieves the highest accuracy (64.94\%), possibly due to its robustness to noisy and unconstrained environments.
    \item EfficientNetV2 is a recent lightweight architecture that incorporates compound scaling and advanced training optimizations. It achieves the best performance on both CASME-II (70.00\%) and SAMM (68.89\%), highlighting its superior capacity to extract discriminative features from phase-specific DIs. Its consistent results across datasets suggest that it offers a strong balance between efficiency and representational power.
    \item ConvNeXt adapts design elements from vision transformers into a CNN framework. Although it has shown promising results in large-scale classification tasks, its relatively lower performance here (e.g., 59.17\% on CASME-II) indicates that deeper and heavier models may not always generalize well in low-data, fine-grained settings such as MER.
    \item MobileViT combines the strengths of CNNs and transformers in a compact form, making it attractive for lightweight applications. However, its performance lags behind other models across all datasets, possibly due to underfitting or difficulties in learning temporal-localized features from DIs.
\end{itemize}

Overall, EfficientNetV2 provides the best trade-off between accuracy and efficiency. Its strong performance on both controlled (CASME-II, SAMM) and in-the-wild (MMEW) datasets suggests it is better suited for capturing the nuanced spatiotemporal features present in dynamic images of micro-expressions. In contrast, larger or transformer-based backbones like ConvNeXt and MobileViT may require more data or different training strategies to be effective in this domain. These findings support the use of modern, lightweight CNNs as backbones in MER, especially when paired with task-specific representations like phase-separated DIs.

\subsection{Attention block selection for DIANet}

To assess the effectiveness of our proposed Cross Attention Fusion Block, we compare it with a simplified variant named Simple Attention Block, which uses a basic dot-product attention and residual fusion.

As shown in Table~\ref{tab:attention}, replacing the Cross Attention Fusion Block with Simple Attention Block results in a notable performance drop across all datasets: from 70.00\% to 66.42\% on CASME-II, from 68.89\% to 63.16\% on SAMM, and from 64.24\% to 50.08\% on MMEW. The performance degradation is especially pronounced on MMEW, which contains greater variability in pose and lighting. This suggests that simple attention fails to capture the nuanced interactions between onset–apex and apex–offset streams under challenging conditions. In contrast, the Cross Attention Fusion Block allows each stream to dynamically attend to salient features in the other via learnable query-key-value projections, promoting rich bidirectional interaction and context-aware fusion. This leads to more discriminative and phase-sensitive feature representations, which are critical in recognizing subtle micro-expressions.

These findings demonstrate that while simple attention mechanisms offer computational simplicity, they are insufficient for modeling the asymmetric and complementary dynamics of micro-expression phases. The proposed cross-attention strategy, albeit slightly more complex, substantially improves recognition performance and justifies its integration into DIANet.

\begin{table*}[h]
\caption{Comparison of attention block performance (accuracy \%) on DIANet across three datasets}
    \centering
    \begin{tabular}{cccc}
    \hline
    \toprule
      \textbf{Attention} &\textbf{ CASME-II} & \textbf{SAMM} & \textbf{MMEW} \\
      \midrule
        Simple Attention Block & 66.42 & 63.16 & 50.08 \\
       \textbf{ Cross Attention Fusion Block (ours)} & \textbf{70.00} & \textbf{68.89} & \textbf{64.24} \\
        \bottomrule      
    \end{tabular}

    \label{tab:attention}
\end{table*}

\subsubsection{Effect of phase-wise dynamic images}
To assess the contribution of phase-specific motion modeling, we conduct an ablation study comparing different input configurations within the DIANet framework, as summarized in Table~\ref{tab:dual-stream}. Specifically, we evaluate three types of inputs: (1) standard dynamic images generated from the entire sequence (as used in LEARNet~\cite{verma2019learnet}), (2) dynamic images computed separately for each temporal phase (DI-Onset and DI-Offset), and (3) a dual-stream setting where both DI-Onset and DI-Offset are processed in parallel.

\begin{table*}[ht]
\centering
\renewcommand{\arraystretch}{1.3}
\caption{Performance comparison of phase-wise dynamic image on three datasets (accuracy \%)}
\label{tab:dual-stream}
\begin{tabular}{llcccc}
\toprule
 & \textbf{Input} & \textbf{Method} & \textbf{CASME-II} & \textbf{SAMM} & \textbf{MMEW} \\
\midrule
\multirow{3}{*}{1-stream} 
  & Dynamic image           & LearNet*~\cite{verma2019learnet} & 56.67 & 59.26 & 48.96 \\
  & \textbf{DI-Onset}       & LearNet*~\cite{verma2019learnet} & \textbf{60.00} & 54.07 & \textbf{52.78} \\
  & \textbf{DI-Offset}      & LearNet*~\cite{verma2019learnet} & 57.08 & \textbf{61.48 }& 50.35 \\
\midrule
\multirow{2}{*}{2-stream} 
  & Dynamic image     & DIANet    & 58.33 & 59.26 & 51.04 \\
  & \textbf{DI-Onset + DI-Offset} & \textbf{DIANet} & \textbf{70.00} & \textbf{68.89} & \textbf{64.24} \\
\bottomrule
\end{tabular}

\vspace{5pt} 
\small{ * This results are from our re-implementation with the model provided by the author}
\end{table*}

In the single-stream setting, using phase-specific DIs (either DI-Onset or DI-Offset) yields better performance than using a holistic dynamic image. For instance, on CASME-II, DI-Onset improves accuracy from 56.67\% (standard DI) to 60.00\%, while on MMEW, it raises accuracy from 48.96\% to 52.78\%. Similarly, DI-Offset provides the best single-phase result on SAMM (61.48\%), outperforming both the standard DI and DI-Onset. These improvements indicate that motion restricted to a specific phase, either the rising (onset–apex) or falling (apex–offset) interval, contains more discriminative features for recognizing micro-expressions than aggregating motion over the entire sequence. This supports our initial hypothesis that the dynamics of MEs follow an asymmetric temporal pattern, and treating the sequence holistically may dilute critical motion cues.

In the dual-stream configuration, DIANet processes both DI-Onset and DI-Offset simultaneously, with a cross-attention fusion module learning to integrate complementary information from the two phases. This setup achieves the best performance across all datasets: 70.00\% on CASME-II, 68.89\% on SAMM, and 64.24\% on MMEW. Notably, the dual-stream DIANet outperforms all single-stream variants, including both phase-specific and standard DI inputs.

These results provide strong empirical evidence for the effectiveness of phase-aware modeling in MER. The dual-stream input not only leverages the unique temporal properties of each phase but also enables the model to learn richer and more balanced representations of facial motion. The significant performance gap between the single-phase and dual-phase settings further validates the design choice of using DI-Onset and DI-Offset as complementary inputs within the DIANet architecture.

\section{Conclusion}
\label{s: conclusion}
This paper has proposed DIANet, the first dual-stream micro-expression recognition framework that leverages phase-aware dynamic images to model facial motion with greater temporal precision. By explicitly separating the onset–apex and apex–offset phases, the proposed approach captures complementary motion dynamics that are often overlooked in holistic or single-phase representations. The two streams are integrated through a cross-attention mechanism and guided by a consistency objective, enabling the network to learn more discriminative and temporally aligned features. Extensive experiments conducted on three benchmark datasets (i.e., CASME-II, SAMM, and MMEW) demonstrate the effectiveness of the proposed method. DIANet achieves 70.00\%, 68.89\%, and 64.24\% accuracy on these datasets, respectively, outperforming or matching state-of-the-art approaches without requiring auxiliary modalities or complex fusion schemes. These results underscore the importance of phase-aware modeling in micro-expression recognition. The proposed framework offers a lightweight yet powerful solution that generalizes well across both controlled and unconstrained settings. Future work may explore extending phase-aware modeling to other motion representations and incorporating temporal uncertainty in apex estimation. 

\section*{Acknowledgement}
This work has been supported by HORIZON-MSCA-SE-2022 PhySU-Net 241 project ACMod (grant 101130271).

\bibliography{sample-ceur}

\appendix

\end{document}